\PassOptionsToPackage{numbers, compress}{natbib}
\documentclass{article}

 \usepackage[preprint]{neurips_2026}

\usepackage[numbers, compress]{natbib} 

\usepackage[utf8]{inputenc} 
\usepackage[T1]{fontenc}    
\usepackage{hyperref}       
\usepackage{url}            
\usepackage{booktabs}       
\usepackage{amsfonts}       
\usepackage{nicefrac}       
\usepackage{microtype}      
\usepackage{xcolor}         

\usepackage{amsmath}        
\usepackage{algorithm}
\usepackage{algorithmic}
\usepackage{cleveref}       
\usepackage{graphicx}       
\usepackage{subcaption}     

\usepackage{enumitem}
\usepackage[most]{tcolorbox}

\title{Training Language Agents to Learn from Experience}

\author{%
  Yuval Shalev \\
  University of Cambridge\\
  \texttt{ys716@cam.ac.uk}
  \And
  Zifeng Ding \\
  University of Cambridge\\
  \texttt{zd320@cam.ac.uk}
  \And
  Mateja Jamnik \\
  University of Cambridge\\
  \texttt{mj201@cam.ac.uk}
}

\begin{document}

\maketitle

\begin{abstract}

Language agents can adapt from experience in interactive environments, but current reflection-based methods can only self-correct within a single task instance. Whether such experience can be distilled into reusable lessons that improve performance on future unseen tasks remains unclear.
We address this problem by introducing the In-context Training (ICT) task, a framework for evaluating cross-task self-improvement in language agents. In ICT, a reflector model observes trajectories collected by an actor model and generates system prompts intended to improve the actor's performance on future unseen tasks. We then propose an RL-based training pipeline for learning such reflections directly from experience, without human-provided examples.
Across ALFWorld and MiniHack, our trained reflectors outperform an untrained baseline on most held-out task families, showing that the ability to learn from experience can itself be learned. In some cases, we observe generalisation beyond the benchmark on which the reflector was trained, to substantially different environments. Finally, we introduce MetaGym, a generic Python library for constructing meta-environments, enabling future research on self-improving language agents.\looseness-1

\end{abstract}

\section{Introduction}

In recent years, there has been a growing interest in using large language models (LLMs) as decision-making agents in interactive environments~\citep{wang2024survey,sumers2023cognitive}.
In this setting, LLMs act as agents that leverage their language capabilities to interact with external environments, producing tokens that are interpreted as actions~\citep{yao2022react}.
When faced with an unfamiliar environment, a general-purpose agent should be able to learn and adapt its behaviour from experience in order to find an optimal policy for action~\citep{silver2025welcome, sawyer2024can,ying2025assessing,shinn2023reflexion}.
Humans handle this task with ease. For example, when playing chess for the first time, a person may begin with an initial strategy in mind and, after each game, reflect on their experience to refine that strategy for future matches.

In LLM-based agents, the agent's policy is determined not only by the model's parameters, but also by the prompt provided to the model, which specifies how the agent should act in the environment~\citep{yao2022react,brown2020language}. 
Learning and adapting may therefore be achieved by modifying the prompt provided to the model, making prompt optimisation a natural mechanism for self-learning in language agents.
This approach has three main advantages:
(1)~it avoids the compute-heavy pipeline of fine-tuning LLMs for each specific task through gradient descent;
(2)~it leverages the linguistic abilities of the model to support rapid learning from only a small number of experiences;
and (3)~it offers a degree of transparency, since the accumulated experience is represented solely in natural language and can be directly inspected.\looseness-1

Recent studies have explored self-improvement in language agents through autonomous prompt modification, using self-reflection mechanisms~\citep{shinn2023reflexion,yao2023retroformer,feng2025improving,ge2025samule,jiang2026metarl}. In this setting, an LLM generates a reflection following an unsuccessful attempt at an interactive task. The reflection considers both the agent's actions and the resulting feedback from the environment, and is then used to guide the agent's next attempt by providing hints and suggesting how to avoid similar mistakes.
This yields a continuous online-learning mechanism, in which each task attempt provides an additional data point for improving behaviour after only a small number of trials.

However, existing work primarily focuses on self-correction by repeatedly attempting the same task until success. This introduces several important limitations.
First, the agent acquires experience only within a single task instance, and the resulting knowledge is not transferred across tasks. Once a task has been solved, the agent must repeat the full reflection process from scratch when encountering a new task.
This is inherently inefficient, since insights from one task could often generalise to similar tasks (e.g., \textit{``in maze-solving tasks, avoid revisiting the same path''}). Under the current paradigm, however, the agent is forced to relearn such lessons for each task.
Second, in many real-world settings, repeating the exact same task is not possible. For example, in a game of chess one cannot require an opponent to reproduce the same sequence of moves, and in a cooking scenario a mistake may irreversibly consume the available ingredients.

Although the ability to learn online from experience across tasks is highly desirable, it remains unclear whether self-reflection mechanisms can produce reflections that generalise well enough  to improve performance on previously unseen tasks.
We therefore ask the following research question: \textit{\textbf{Can LLMs be trained to generate self-reflections that facilitate cross-task self-improvement, and if so, how?}}.\looseness-1

\begin{figure}[t] 
  \centering
  \includegraphics[width=0.95\textwidth]{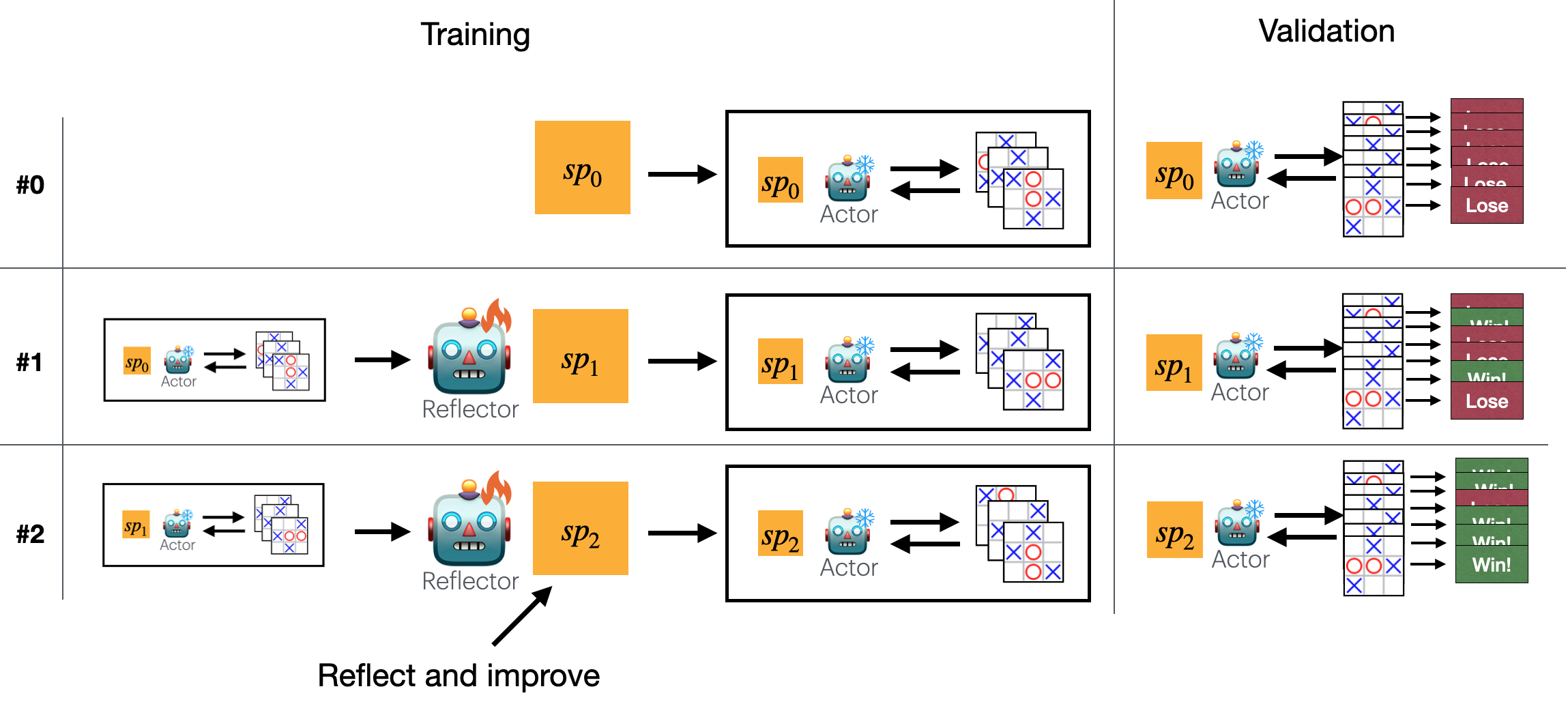}
  \caption{The In-context Training (ICT) task. In each turn, the reflector model generates a new system prompt $sp_i$ based on the previous performance of the actor model. The actor then interacts with a batch of \textbf{unseen} tasks using the newly generated prompt. The quality of the prompt is subsequently evaluated on a fixed held-out set of tasks.
  In our experiments, we fine-tune the reflector while keeping the actor model frozen.} 
  \label{fig:ict_overview}
\end{figure}
To address this, we first define the \textbf{In-context Training} (ICT) task, a framework for evaluating the ability of LLMs to learn from experience for cross-task self-improvement (see \Cref{fig:ict_overview}).
The ICT task implements a simple self-learning loop through a dual-LLM design.
The first model is the \textit{actor}, which serves as the operational agent interacting with external environments according to a system prompt.
Crucially, at each turn of the ICT task, the actor faces a new \textbf{unseen} batch of tasks and has only \textbf{one chance} to solve each task.
The second model is the \textit{reflector}, which evaluates the actor's performance and proposes an improved system prompt for the the next turn.
The goal of the reflector is to learn from the actor's collected experiences in order to generate a system prompt that generalises across tasks within a given family of tasks.
For example, in a maze-solving task, the reflector should learn from previously encountered mazes in order to produce a system prompt that enables the actor to solve \textbf{any} unseen maze drawn from the same task family in a single attempt.\looseness-1

Next, we demonstrate that the ability to learn from experience is itself a learnable skill by introducing a fine-tuning pipeline to train an LLM to generate reflections that facilitate cross-task self-improvement.
We train a \textit{reflector} to analyse a given system prompt alongside experiences collected under that prompt, and to generate an improved system prompt for future unseen tasks. Our method does not rely on human-provided examples.

We evaluate our approach on ALFWorld~\citep{ALFWorld20} and MiniHack~\citep{samvelyan2021minihack}, two benchmarks comprising diverse interactive environments. Using these benchmarks within the ICT framework, we show that a reflector trained on a set of task types can generalise and improve performance on previously unseen task types using only a small number of gathered experiences.
We also show that, in some cases, the model generalises beyond the benchmark on which it was trained, transferring to substantially different environments.

Finally, we introduce MetaGym, a generic Python library for constructing gym-like meta-environments from any gym-like environment~\citep{1606.01540}.
We release MetaGym as a tool to support future research on self-improvement in LLM agents within the ICT framework and beyond (\url{https://github.com/yuvalshalev1/MetaGym}).

Our contributions can be summarised as follows:
\begin{enumerate}[topsep=0pt, leftmargin=12pt, itemsep=0pt]
  \item A {\bf training pipeline} for learning reflection strategies for cross-task self-improvement without reliance on human-provided labels.
  \item The \textbf{In-context Training} task, a framework for evaluating online self-improvement of LLMs that generalises across task instances.
  \item {\bf MetaGym}, a Python library for constructing meta-environments from Gym-like environments.
\end{enumerate}

\section{Related Work}
\textbf{Prompt Optimisation.}
Previous studies have demonstrated the potential of LLMs for prompt optimisation in static tasks \cite{yang2023large,choi2025system,agrawal2025gepa}.
\citet{yang2023large} optimise prompts for mathematical reasoning and question answering using closed-source LLMs, and \citet{choi2025system} propose a bi-level framework that jointly optimises task-specific user prompts and a cross-task system prompt. In both cases, the optimiser observes only final outputs or aggregate accuracy, not \textbf{how} its prompts shaped the model's behaviour during problem solving.
In our setting, however, the optimiser observes full interaction trajectories, enabling us to study its ability to learn from the agent's behaviour.
\citet{agrawal2025gepa} leverage full reasoning trajectories to optimise prompts through iterative updates to a prompt pool. However, their scoring procedure relies on evaluating multiple candidate prompts on the same batch of task instances. In contrast, our setting assumes that each task is encountered only once and learning must occur online from non-repeatable experience.
More broadly, these works optimise prompts using external procedure rather than fine-tuning the optimiser itself, leaving open whether LLMs can be trained to learn better optimisation strategies.

\textbf{Prompt Optimisation for LLM-Agents.}
Recent work has explored prompt optimisation for LLM agents in interactive environments, while transferring experience across  tasks~\citep{sarukkai2025self,zhao2024expel}.
\citet{sarukkai2025self} maintain an external database of self-generated successful trajectories and retrieve similar examples as in-context demonstrations for the actor.
This method relies on embedding raw past trajectories directly into the prompt, rather than consolidating them into a structured representation. However, prior work shows that in-context demonstrations are not always effective, even when produced by human experts~\citep{ruoss2024lmact}.
\citet{zhao2024expel} similarly rely on an external mechanism, extracting “insights” from pairs of past experiences and maintaining them through a  voting procedure, before combining them with selected raw experiences to construct task-specific prompts during inference time. These methods therefore transfer experience through external memory and retrieval, but leave open the question of whether an LLM can be trained to learn reusable reflective strategies for cross-task self-improvement. \looseness-1

\textbf{Reflection for Self-Correction.}
\citet{shinn2023reflexion} introduce reflection for LLM agents as a feedback loop in which an agent analyses its behaviour after failing a task and uses this analysis to guide subsequent attempts.
Since then, several works have explored how to train LLMs to reflect more effectively in interactive environments. 
\citet{yao2023retroformer,ge2025samule} and \citet{feng2025improving} use dual-LLM architecture that separates actor and reflector, with the former training only the reflector and the latter training both components jointly using a gradient policy.
Most recently, \citet{jiang2026metarl} proposed a meta-RL framework in which a single LLM acts as both actor and reflector.
However, all of these methods focus on self-correction through repeated attempts at the same task. They, therefore, leave open whether reflection can support cross-task improvement by extracting reusable lessons from experience.\looseness-1

\section{The In-context Training Task}\label{sec:ict}
Here, we introduce the In-context Training (ICT) task, a framework for evaluating whether large language models can learn from experience to achieve cross-task self-improvement. 
ICT is designed to capture a realistic setting in which the same task cannot be repeated, and the only opportunity for improvement lies in future tasks.
The framework is intentionally minimalistic, aiming to measure the model's intrinsic learning ability rather than optimise performance through external machinery. 

\begin{algorithm}[t]
  \caption{In-context Training (ICT)}
  \label{alg:ict}
  \begin{algorithmic}[1]
  \REQUIRE $\text{LLM}_\text{refl}$, $\text{LLM}_\text{act}$, $E^\text{train}$, $E^\text{val}$, batch size $k$, meta-turns $N$, initial system prompt $sp_0$
  \STATE $sp^* \leftarrow sp_0$, $\quad score^* \leftarrow 0$
  \STATE Sample $\{e^{(0)}_j\}_{j=1}^{k} \sim E^\text{train}$
  \STATE $obs_0 \leftarrow \bigl\{\tau^{(0)}_j \sim \text{LLM}_\text{act}(e^{(0)}_j;\, sp_0)\bigr\}_{j=1}^{k}$

  \FOR{meta-turn $i = 1, \dots, N$}
      \STATE Generate new system prompt $sp_i \sim \text{LLM}_\text{refl}(sp_{i-1}, obs_{i-1})$

      \STATE $score_i \leftarrow \sum_{e \in E^\text{val}} R(\text{LLM}_\text{act}, e, sp_i)$
      \IF{$score_i > score^*$}
          \STATE $sp^* \leftarrow sp_i$, $\quad score^* \leftarrow score_i$
      \ENDIF

      \STATE Sample $\{e^{(i)}_j\}_{j=1}^{k} \sim E^\text{train}$
      \STATE $obs_i \leftarrow \bigl\{\tau^{(i)}_j \sim \text{LLM}_\text{act}(e^{(i)}_j;\, sp_i)\bigr\}_{j=1}^{k}$
  \ENDFOR

  \RETURN $sp^*$
  \end{algorithmic}
\end{algorithm}

The full process is formulated in \Cref{alg:ict} and illustrated in \Cref{fig:ict_overview}. We first define the reflector model (denoted $\text{LLM}_\text{refl}$), whose learning capabilities we aim to evaluate, the actor model (denoted $\text{LLM}_\text{act}$), whose policy we aim to optimise, and a class of tasks $E$ on which training is performed. The task class is partitioned into a training set $E^\text{train}$ and a validation set $E^\text{val}$.
The objective is for the tested $\text{LLM}_\text{refl}$ to analyse episodes sampled from $E^\text{train}$ using $\text{LLM}_\text{act}$ and generate a system prompt $sp$ that enables the actor $\text{LLM}_\text{act}$ to perform effectively across all tasks in $E^\text{val}$.\looseness-1

Formally, the objective of the reflector $\text{LLM}_\text{refl}$ is to generate a system prompt $sp^*$ that maximises the cumulative reward obtained by the actor across the validation tasks:
\begin{equation}
  sp^* = \operatorname*{arg\,max}_{sp} \sum_{i=1}^{|E^{val}|} R(\text{LLM}_\text{act},e_i,sp)
\end{equation}
where $R(\text{LLM}_\text{act}, e_i, sp)$ denotes the reward obtained by the actor $\text{LLM}_\text{act}$ on task $e_i \in E^{val}$ when conditioned on the system prompt $sp$.

The reflector $\text{LLM}_\text{refl}$ is modelled as a meta-agent interacting with a meta-environment.
The meta-environment is initialised with an initial system prompt $sp_0$ and samples a batch of $k$ tasks from the training set $E^\text{train}$. The meta-environment then executes a batch of episodes using $sp_0$ and returns a meta-observation $obs_0$, the full sequences of interactions between the actor $\text{LLM}_\text{act}$ and the sampled environments (line 3 of \Cref{alg:ict}).
Then, the meta-loop begins: at each meta-turn $i$, the reflector $\text{LLM}_\text{refl}$ is prompted to generate a new system prompt $sp_i$, conditioned on the previous system prompt $sp_{i-1}$ and the last meta-observation $obs_{i-1}$ (line 5):
\begin{equation} \label{eq:sp_i}
  sp_i \sim \text{LLM}_\text{refl}(sp_{i-1}, obs_{i-1})
\end{equation}
where $obs_{i}$ is the meta-observation consisting of the batch of trajectories collected under $sp_{i}$:
\begin{equation} \label{eq:obs_i}
  obs_{i} = \{\tau^{(i)}_j \sim \text{LLM}_\text{act}(e^{(i)}_j, sp_{i})\}_{j=1}^{k}
\end{equation}
and $\tau^{(i)}_j = (a_0, o_0, a_1, o_1, \dots, a_T, o_T)$ is the trajectory of actions
$a_t$ and environment observations $o_t$ produced by $\text{LLM}_\text{act}$ on task $e^{(i)}_j$ under system prompt $sp_i$.

The proposed system prompt $sp_i$ is then evaluated on the validation tasks $E^\text{val}$, and sent to the meta-environment to execute the next batch of episodes (lines 10-11).

\section{Training LLMs to Learn from Experience}\label{sec:training}

We now introduce our pipeline for fine-tuning LLMs to generate reflections that facilitate cross-task self-improvement. 

\textbf{Dataset Creation.}\label{sec:dataset_creation}
To construct the training dataset, we run the ICT loop $r$ times for $N$ turns each, using a base model.
At the end of each turn $i$, we record the tuple $(sp_i, obs_i, e^{(i)}_0, \dots, e^{(i)}_{k-1})$, consisting of the current system prompt $sp_i$, the meta-observation $obs_i$, and the batch of tasks $e^{(i)}_0, \dots, e^{(i)}_{k-1}$ used in that turn. Repeating this process yields a fixed dataset~$\mathcal{D}$ of $r \times N$ data points in total.
The motivation is to capture a diverse set of system prompts arising from different stages of the ICT process, which correspond to varying degrees of required improvement. For example, if the reflector receives a system prompt that succeeds on one task in the batch but fails on the others, it should be able to analyse the factors that contributed to the successful case and generalise them to the remaining tasks. Conversely, if the reflector receives a system prompt that succeeds on all tasks in the batch, it should generate a new system prompt that remains highly similar to the previous one, potentially reproducing it exactly.\looseness-1

\textbf{Reward Signal.}
As described in \Cref{eq:sp_i}, the reflector generates a system prompt $sp_i$ conditioned on the previous system prompt $sp_{i-1}$ and the last batch of trajectories $obs_{i-1}$ collected by $\text{LLM}_\text{act}$ under $sp_{i-1}$.
Its immediate task is therefore to analyse how the previous system prompt led to the actor's observed  performance on the previous batch $e^{(i-1)}$, and to improve the prompt in light of the issues revealed by those trajectories.
Put differently, although the reflector's goal is to improve performance on future unseen tasks, its intermediate objective is to answer the following question: \textit{``If I had the opportunity to try the same batch of tasks again, what would I do differently to improve my performance?''}.\looseness-1

To encourage the model to account for observed performance issues and incorporate corresponding fixes into the new system prompt, we derive the training reward directly from the previous batch of tasks.
Specifically, the reward used to rank $sp_i$ is the average reward obtained by the actor $\text{LLM}_\text{act}$ on the batch $e^{(i-1)}$ when conditioned on $sp_i$, that is, by replaying the previous batch under the new system prompt $sp_i$. Formally, the reward function is defined as:
\begin{equation} \label{eq:reward}
  \mathcal{R}(sp_i) = \frac{1}{k} \sum_{j=1}^{k} R(\text{LLM}_\text{act},\, e^{(i-1)}_j,\, sp_i).
\end{equation}
Notably, although the reward signal is derived from the previous batch of tasks, the reflector should be explicitly prompted to analyse those performance issues and generate a system prompt intended for \textbf{future} tasks. This is essential to discourage in-context overfitting to the most recently observed batch. \looseness-1

\begin{figure}[!t]
  \centering
  \includegraphics[width=0.95\textwidth]{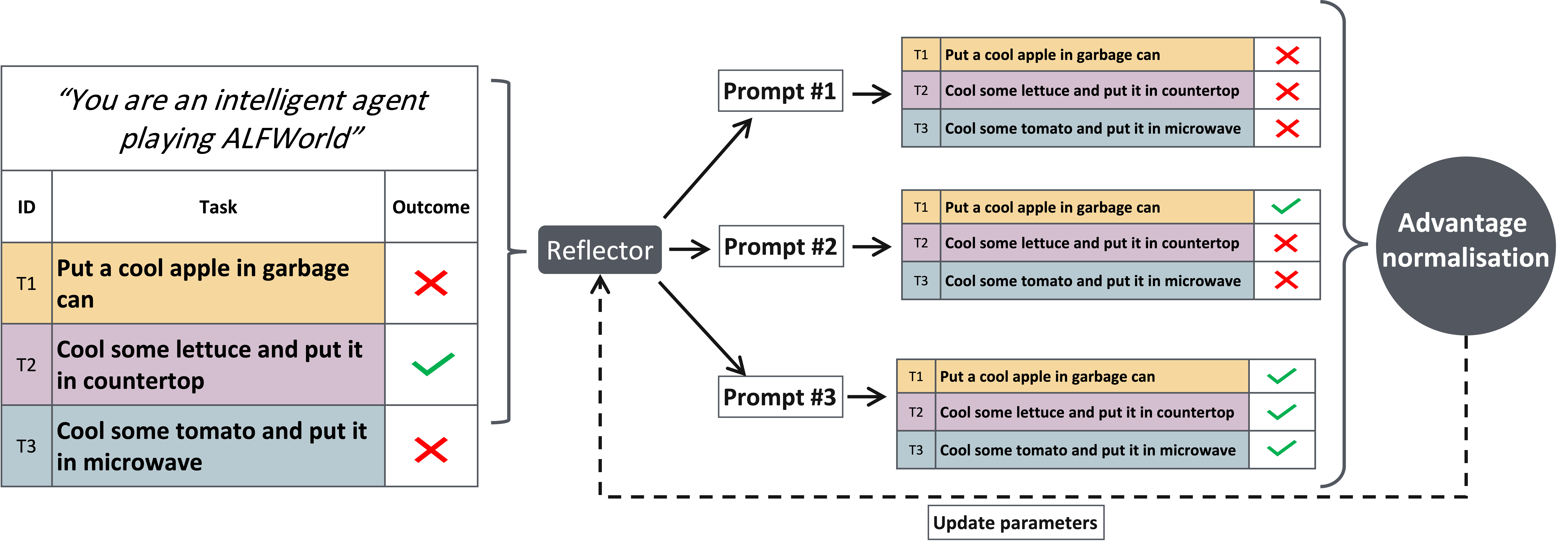}
  \caption{A single training step illustrated on an \textit{ALFWorld-Cool and Place} example. A system prompt and its associated episode batch are sampled from the dataset and fed into the reflector, which generates multiple candidate system prompts. Each candidate is scored by replaying the \textbf{same} tasks with the frozen actor, and the resulting scores drive advantage estimation and parameter updates via GRPO.\looseness-1}
  \label{fig:training_procedure}
\end{figure}

\textbf{Training Procedure.}
We follow a standard training procedure for RL-based fine-tuning, where in each training step, we sample a batch of tuples $(sp_i, obs_i, e^{(i)}_0, \dots, e^{(i)}_{k-1})$ from the dataset~$\mathcal{D}$ and update the parameters of $\text{LLM}_\text{refl}$ using GRPO~\citep{shao2024deepseekmath}.
The procedure is illustrated in \Cref{fig:training_procedure}:
for each tuple, the reflector samples a group of $G$ candidate system prompts 
$\{sp^{(1)}, \dots, sp^{(G)}\} \sim \pi_\theta(\cdot \mid sp_i, obs_i)$, 
and each candidate is scored using $\mathcal{R}$ as defined in \Cref{eq:reward}. 
The scores are computed dynamically during training by executing the batch of tasks
$e^{(i)}_0, \dots, e^{(i)}_{k-1}$ with $\text{LLM}_\text{act}$ under each candidate system prompt.\looseness-1

\section{Experiments}\label{sec:experiments}
\subsection{Experimental Setup}\label{sec:experimental_setup}
All our experiments were conducted using the Qwen2.5-7B-Instruct model~\citep{qwen2} and the \textit{verl} framework~\citep{sheng2024hybridflow} for RL-based fine-tuning.
The datasets were constructed as described in \Cref{sec:dataset_creation}, using $r=128$ ICT loops of $N=4$ turns each, with Qwen2.5-7B-Instruct serving as both the actor model and the base reflector model.
For each benchmark, we trained two reflectors: one with batch size $k=3$ and another with batch size $k=5$.
The reflection prompts used in our experiment are detailed in \Cref{appendix:prompting}.
Training hyperparameters and compute resources are reported in \Cref{appendix:hyperparameters}.

\textbf{MetaGym.} Task execution for both the ICT loop and the fine-tuning procedure is managed by MetaGym, our library for meta-environment construction for LLM-agentic tasks. MetaGym instantiates an interactive meta-environment in which each meta-action is represented by a system prompt, and each returned meta-observation is a batch of rollouts generated by an actor. Internally, it spawns task instances and executes them in parallel. See \Cref{sec:metagym} for more details and a usage example. \looseness-1

\textbf{Task Partitioning.}
The objective of the proposed training approach is to train a reflector that performs effectively on the ICT task, where the goal is to generate a general system prompt that achieves strong performance across tasks within a given task type.
A straightforward evaluation strategy would be to divide task instances into training and validation sets. However, this is insufficient: a reflector trained on one task type could simply memorise a high-performing prompt for that type and reproduce it at test time, thereby bypassing the in-context reasoning the model is intended to learn.
To mitigate this, we partition tasks by \textbf{type} rather than by instance, yielding disjoint \textbf{meta-train} and \textbf{meta-test} sets, each comprising distinct task types.
Each task type consists of a collection of similar tasks, from which the reflector is expected to extract transferable insights that generalise to other tasks within the same type.
During training, the reflector is exposed exclusively to task types from the meta-train set. At evaluation time, it is required to generalise to previously unseen task types in the meta-test set, generating reflections based solely on in-context observations available at inference time. \looseness-1

\subsubsection{Environments}

To construct the meta-train and meta-test sets, we use agentic benchmark environments that could be partitioned into distinct task types.
Specifically, we use ALFWorld~\citep{ALFWorld20} and a subset of MiniHack~\citep{samvelyan2021minihack}.
For each benchmark, we train a dedicated reflector on the tasks in the corresponding meta-train set, and then evaluate it on tasks drawn from the meta-test set of the same benchmark.

\textbf{ALFWorld.}
ALFWorld is a benchmark comprising six task types, all set in a partially observable, house-like textual environment.
Following \citet{jiang2026metarl}, we use \textit{Pick and Place}, \textit{Examine in Light}, \textit{Clean and Place}, and \textit{Heat and Place} as the meta-train set, and \textit{Cool and Place} and \textit{Pick Two and Place} as the meta-test set.
For dataset creation, we execute the ICT loop 32 times for each of the four meta-train task types. Further details are provided in \Cref{appendix:alfworld}.

\textbf{MiniHack.}
MiniHack is a benchmark comprising many tasks in two-dimensional grid world environments.
To construct a meta-set with different task types, we use a collection of eight 5x5 grid world environments.
We use \textit{MiniHack-Room-Random-5x5-v0}, \textit{MiniHack-Eat-v0}, \textit{MiniHack-Wield-v0}, and \textit{MiniHack-Wear-v0} as the meta-train set, and \textit{MiniHack-PutOn-v0}, \textit{MiniHack-Zap-v0}, \textit{MiniHack-Read-v0}, and \textit{MiniHack-Room-Dark-5x5-v0} as the meta-test set.
For dataset creation, we execute the ICT loop 32 times for each of the four meta-train task types. Further details are provided in \Cref{appendix:minihack}. \looseness-1


\subsection{Results}
We evaluate the reflectors trained on ALFWorld and MiniHack on their respective held-out meta-test sets.
The ALFWorld reflectors are evaluated using the \textit{Cool and Place} and \textit{Pick Two} task types, and the MiniHack reflectors were evaluated on the \textit{Dark}, \textit{Read}, \textit{PutOn}, and \textit{Zap} task types.
Each reflector is tested using the same batch size it is trained on.
For each reflector and task type, we run the ICT task 10 times for 10 turns each. 
This repetition is designed to control for noise introduced by model temperature and to evaluate robustness across different in-context training tasks.
Each run is initialised using a different random seed to obtain different task batches, while the validation sets are kept fixed across all runs.
For each run, we select the best-performing prompt, and the average success rates across the winning prompts are summarised in \Cref{tab:reflector_comparison}, along with the performance of the untrained Qwen2.5-7-Instruct model and the average result achieved using the initial prompt.
For calculating the average result of the initial prompt, we use the initial outcomes from all runs across all four reflectors ($n = 40$).
\Cref{fig:examples} shows representative examples of reflector outputs on the \textit{MiniHack-Read} and \textit{ALFWorld Cool and Place} tasks.
Below we discuss our findings. \looseness-1

\begin{table}[!t]
   \caption{Average winning prompt success rates of our trained reflectors viz.\ Qwen2.5-7B-Instruct and the initial prompt ($sp_0$), on held-out task types for varying batch sizes $k$. $\pm$ values indicate standard deviation across all random seeds. 
   Our trained reflector shows better performance on the ICT task across nearly all task types in both benchmarks.}
  \label{tab:reflector_comparison}
 \label{tab:reflector_comparison}
  \centering
  \resizebox{\textwidth}{!}{%
  \begin{tabular}{lccccc}
  \toprule
  & & \multicolumn{2}{c}{\textbf{$k=3$}} & \multicolumn{2}{c}{\textbf{$k=5$}} \\
  \cmidrule(lr){3-4} \cmidrule(lr){5-6}
  \textbf{Environment} & \textbf{Initial prompt} & \textbf{Qwen2.5-7B} & \textbf{Trained (ours)} & \textbf{Qwen2.5-7B} & \textbf{Trained (ours)} \\
  \midrule
  \multicolumn{6}{l}{\textit{ALFWorld}} \\
  \quad Cool and Place  & 13.3{\scriptsize$\pm$4.3}\% & 42.4{\scriptsize$\pm$10.9}\% & \textbf{49.2{\scriptsize$\pm$10.1}\%} & 38.8{\scriptsize$\pm$12.0}\% & \textbf{48.0{\scriptsize$\pm$15.0}\%} \\
  \quad Pick Two        & 27.7{\scriptsize$\pm$6.5}\% & 34.6{\scriptsize$\pm$5.6}\% & \textbf{42.5{\scriptsize$\pm$7.2}\%} & 28.3{\scriptsize$\pm$6.1}\% & \textbf{41.2{\scriptsize$\pm$7.6}\%} \\
  \midrule
  \multicolumn{6}{l}{\textit{MiniHack}} \\
  \quad Dark            & 35.0{\scriptsize$\pm$7.2}\% & 48.4{\scriptsize$\pm$5.8}\% & \textbf{51.9{\scriptsize$\pm$4.0}\%} & 42.8{\scriptsize$\pm$9.7}\% & \textbf{48.1{\scriptsize$\pm$3.5}\%} \\
  \quad Read            & 29.1{\scriptsize$\pm$6.4}\% & 34.7{\scriptsize$\pm$4.1}\% & \textbf{38.4{\scriptsize$\pm$5.4}\%} & 32.2{\scriptsize$\pm$5.9}\% & \textbf{40{\scriptsize$\pm$5.4}\%} \\
  \quad PutOn           & 24.8{\scriptsize$\pm$5.2}\% & 26.6{\scriptsize$\pm$9.0}\% & \textbf{30{\scriptsize$\pm$4.7}\%} & 22.8{\scriptsize$\pm$10.0}\% & \textbf{34.1{\scriptsize$\pm$2.2}\%} \\
  \quad Zap             & 10.4{\scriptsize$\pm$5.8}\% & \textbf{18.8{\scriptsize$\pm$5.6}\%} & 14.1{\scriptsize$\pm$3.5}\% & 23.1{\scriptsize$\pm$7.9}\% & \textbf{27.8{\scriptsize$\pm$4.5}\%} \\
  \bottomrule
  \end{tabular}}
  \vspace*{-3mm}
\end{table}

\begin{figure}[!t]
\begin{tcolorbox}[
  title=MiniHack - Read,
  colback=gray!5,
  colframe=gray!50,
  fonttitle=\bfseries,
  fontupper=\small
]
\textbf{=== ANALYSIS ===}

In Episode~1, the agent \textbf{successfully picks up and reads} a resource. The system prompt did not specify what to do in this situation, and the agent made the correct decisions.

In Episodes~2 and~3, the agent either makes incorrect assumptions or takes actions that lead to a negative reward due to misread or misinterpreted situations.

The system prompt needs to clearly guide the agent on what to do in all situations [\ldots]

\smallskip
\textbf{=== IMPROVED PROMPT ===}

You are an intelligent agent playing [\ldots]

Continue exploring until you either find a resource or item, or until you have explored all possible directions. [\ldots]

If you find a resource or item, pick it up or \textbf{read it}, depending on the situation.
\end{tcolorbox}
\vspace{1pt}
\begin{tcolorbox}[
  title=ALFWorld - Cool and Place,
  colback=gray!5,
  colframe=gray!50,
  fonttitle=\bfseries,
  fontupper=\small
]
\textbf{=== ANALYSIS ===}

In all three episodes, the agent failed to correctly identify the location of a cool potato or a plate. [\ldots]

For improvement, \textbf{the prompt should more strongly guide the agent to consider locations like fridges} and side tables when an object needs to be cooled. The prompt should ensure the agent knows to go to these locations first.

\smallskip
\textbf{=== IMPROVED PROMPT ===}

You are an intelligent agent playing [\ldots]

If you see an object you need to cool, \textbf{consider locations such as the fridge} and side tables, and attempt to go there first to pick up the object and cool it more quickly. [\ldots]

Remember to safely handle the cooled object and place it in the required location after cooling.
\end{tcolorbox}
\caption{Examples of our reflectors' output on held-out task types.
\textbf{(Top)} The reflector correctly deduces that the goal of \textit{MiniHack-Read} environments is to find the item and read it, and generates a system prompt that guides the agent to do so.
\textbf{(Bottom)} The reflector observes that the actor is struggling to understand how to cool the objects in the \textit{ALFWorld Cool and Place} environments, and generates a system prompt that guides the agent to look for the fridge.
There were no tasks requiring cooling or reading in the training data.}
\label{fig:examples}
\end{figure}

\textbf{LLMs Can Be Trained to Achieve Cross-Task Self-Improvement.}
\Cref{tab:reflector_comparison} reveals that across nearly all evaluated tasks and environments, our trained reflector consistently outperforms the untrained Qwen2.5-7B-Instruct baseline, demonstrating that the ability to learn from experience is itself a learnable capability.
Reflectors trained on one set of task types successfully transfer their improved reflection ability to unseen tasks types -- providing direct evidence that \textit{learning from experience} is itself a learnable meta-skill.
The one exception is the \textit{Zap} task in MiniHack for the $k{=}3$ condition, which is the hardest task and may require larger batches to meaningfully learn from.
Notably, our results indicate that untrained Qwen2.5-7B-Instruct was also able to improve the initial system prompt when conditioned on sampled episodes of the actor, suggesting that the ability to extract reusable lessons from experience already exists in the model to some extent.

\begin{figure}[!t]
    \centering
    \begin{subfigure}[b]{0.48\textwidth}
        \includegraphics[width=\textwidth]{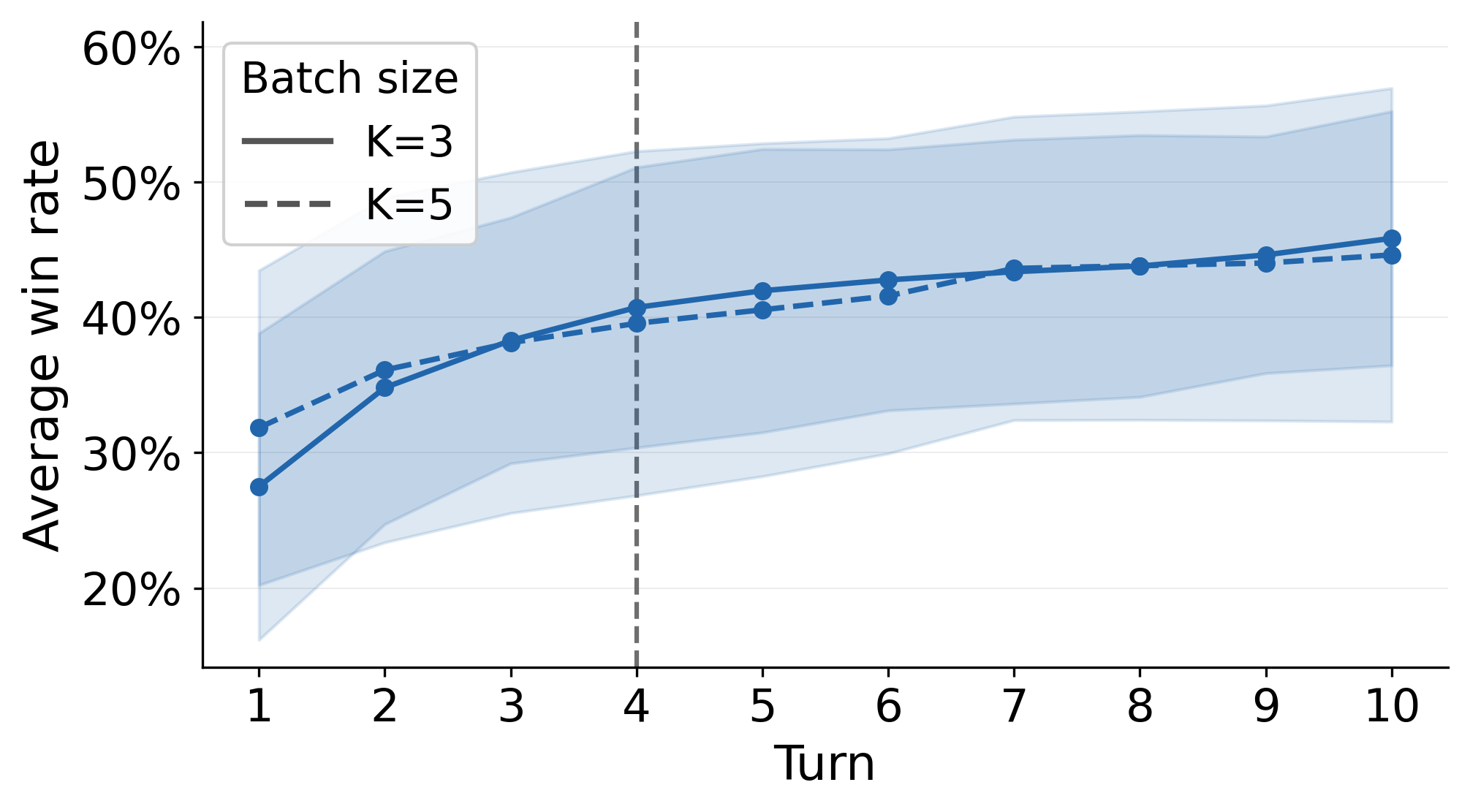}
        \caption{ALFWorld}
        \label{fig:avg_best_over_turns}
    \end{subfigure}
    \hfill
    \begin{subfigure}[b]{0.48\textwidth}
        \includegraphics[width=\textwidth]{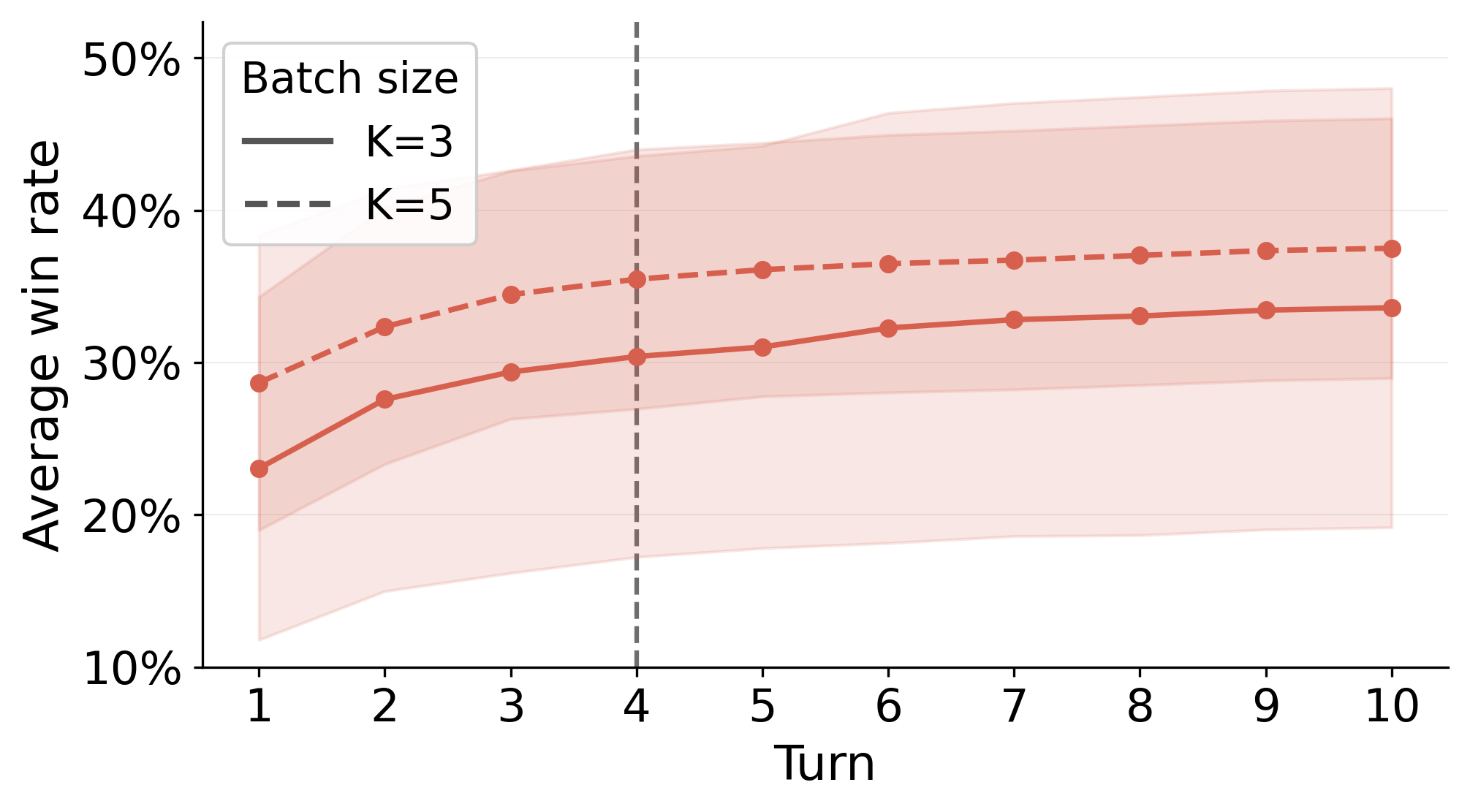}
        \caption{MiniHack}
        \label{fig:avg_best_over_envs}
    \end{subfigure}
    \caption{Average winning prompt success rate up to each turn, aggregated across all tasks for each reflector. Shaded regions indicate standard deviation across all random seeds. The dashed vertical line marks turn 4, the end of the training horizon. Performance continues to improve beyond the number of steps seen during training.}
    \label{fig:avg_best_groups}
    \vspace{-10pt}
\end{figure}

\textbf{More Meta-Turns Lead to Better Performance.}
\Cref{fig:avg_best_groups} presents the average best win rate achieved up to each turn, aggregated across all tasks for each experimental group.
Performance improves consistently with additional meta-turns across all four groups and, crucially, continues to increase beyond turn 4, the boundary of our training data.
This is an important finding: during training, our model was exposed to prompts created only by the first 4 turns of the ICT loop, yet at test time, it was able to recursively improve its own generated prompts further beyond this horizon.
In the ALFWorld experiments, 55\% ($k{=}3$) and 45\% ($k{=}5$) of the best-performing prompts were achieved after turn 4, while in the MiniHack experiments, 32.5\% ($k{=}3$) and 37.5\% ($k{=}5$) emerged after turn~4.
Notably, 25\% of the best ALFWorld $k{=}3$ prompts were not attained until turn~10, suggesting that further meta-turns may yield additional improvements.

\textbf{Effect of Batch Size.}
The batch size plays two important roles that can influence the performance of the reflector.
First, it determines the number of episodes observed by the reflector in each turn, thereby affecting the diversity of experiences from which it can learn.
Notably, when a task is difficult, a small batch size may contain no successful episodes, making it more challenging for the reflector to extract useful lessons in the absence of positive examples.
Second, batch size affects the fine-tuning process of the reflector itself:
during training, if none of the generated candidate prompts is successful on any of the replayed tasks, the GRPO algorithm cannot compute advantages because all rewards are zero.
A larger batch size increases the likelihood of obtaining a positive reward signal, particularly when tasks are difficult and rewards are sparse.

Increasing the reflection batch size from $k{=}3$ to $k{=}5$ shows different effects across the benchmarks (see \Cref{tab:reflector_comparison}  and \Cref{fig:avg_best_groups}).
In ALFWorld, performance is similar across batch sizes, suggesting that the quality of individual reflections remains largely unchanged.
Taking into account the total number of tasks experienced, $k{=}3$ shows a clear advantage, as adding more episodes appears to be redundant while costing more solving attempts.
In MiniHack, the performance of the $k{=}5$ reflector was, in most cases, better than that of the $k{=}3$ reflector, suggesting that using larger batches can be more effective for fine-tuning the reflector. \looseness-1

The untrained Qwen2.5-7B-Instruct performs worse under the $k{=}5$ condition across almost all tasks.
Notably, in the \textit{Pick Two} task, the prompts it discovers show performance remaining close to the initial prompt.
One possible explanation is that processing five episodes simultaneously presents a workload that is too high for Qwen2.5-7B to handle effectively. One interesting exception is the \textit{Zap} task in MiniHack, where Qwen2.5-7B-Instruct performs better in the k=5 condition, which we attribute to the difficulty of the task.

\textbf{Cross-Benchmark Generalisation.}
To test whether the meta-skill learned by our method generalises beyond the training benchmark, we evaluate reflectors trained only on MiniHack within ALFWorld.
Results are shown in \Cref{tab:reflector_comparison_cross}, alongside the initial prompt and Qwen2.5-7B-Instruct baselines.
The reflector trained on MiniHack with a batch size of $k{=}5$ outperforms the untrained Qwen2.5-7B-Instruct baseline on 4 out of 6 tasks, while achieving comparable performance on the remaining two.
Notably, on both the \textit{Examine in the Light} and \textit{Heat and Place} tasks, it outperforms Qwen2.5-7B-Instruct even under the $k{=}3$ condition, which we previously observed in \Cref{tab:reflector_comparison} to be easier within ALFWorld.
This is a significant finding, as it demonstrates transferability between fundamentally different benchmarks.
By contrast, the $k{=}3$ reflector does not achieve improved performance and, in some cases, performs worse than the untrained model.
We attribute this, as discussed earlier,  to more difficult training at smaller batch sizes, where the reward signal is sparser.\looseness-1

\begin{table}[!t]
  \caption{Cross-benchmark evaluation: average winning prompt success rate for MiniHack reflector evaluated on ALFWorld. $\pm$ values indicate standard deviation across all random seeds. 
  Our trained $k=5$ reflector was able to transfer its learning capabilities to substantially different task types.}
  \label{tab:reflector_comparison_cross}
  \centering
  \resizebox{\textwidth}{!}{%
  \begin{tabular}{lcccccc}
  \toprule
  & & \multicolumn{2}{c}{\textbf{$k=3$}} & \multicolumn{2}{c}{\textbf{$k=5$}} \\
  \cmidrule(lr){3-4} \cmidrule(lr){5-6}
  \textbf{Environment} & \textbf{sp$_0$} & \textbf{Qwen2.5-7B} & \textbf{Trained (ours)} & \textbf{Qwen2.5-7B} & \textbf{Trained (ours)} \\
  \midrule
  \quad Pick and Place          & 78.1{\scriptsize$\pm$3.4}\% & \textbf{89.1{\scriptsize$\pm$4.7}\%} & 86.6{\scriptsize$\pm$4.6}\% & \textbf{87.8{\scriptsize$\pm$2.6}\%} & 83.4{\scriptsize$\pm$4.2}\% \\
  \quad Examine in the Light    & 45.2{\scriptsize$\pm$8.3}\% & \textbf{66.2{\scriptsize$\pm$9.2}\%} & 64.6{\scriptsize$\pm$12.0}\% & 61.5{\scriptsize$\pm$8.4}\% & \textbf{73.1{\scriptsize$\pm$8.6}\%} \\
  \quad Clean and Place         & 34.4{\scriptsize$\pm$5.7}\% & \textbf{56.7{\scriptsize$\pm$8.6}\%} & 39.6{\scriptsize$\pm$6.6}\% & 50.7{\scriptsize$\pm$11.1}\% & \textbf{57.8{\scriptsize$\pm$5.8}\%} \\
  \quad Heat and Place          & 16.6{\scriptsize$\pm$8.1}\% & \textbf{25.0{\scriptsize$\pm$8.8}\%} & 19.4{\scriptsize$\pm$7.1}\% & 21.9{\scriptsize$\pm$8.0}\% & \textbf{40.6{\scriptsize$\pm$11.9}\%} \\
  \quad Cool and Place          & 13.3{\scriptsize$\pm$4.1}\% & \textbf{42.4{\scriptsize$\pm$10.9}\%} & 27.6{\scriptsize$\pm$6.6}\% & \textbf{38.8{\scriptsize$\pm$12.0}\%} & 34.0{\scriptsize$\pm$6.0}\% \\
  \quad Pick Two                & 28.2{\scriptsize$\pm$5.2}\% & \textbf{34.6{\scriptsize$\pm$5.6}\%} & 25.4{\scriptsize$\pm$5.1}\% & 28.3{\scriptsize$\pm$6.1}\% & \textbf{35.0{\scriptsize$\pm$5.0}\%} \\
  \bottomrule
  \end{tabular}
  }
\end{table}

\vspace*{-2mm}
\subsection{Limitations}\label{sec:limitations}
\vspace*{-1mm}
In our experiments, we trained only the reflector model in order to isolate the contribution of prompt evolution to actor performance.
However, maintaining separate actor and reflector models is computationally expensive in practice, as it requires holding two models in memory simultaneously. Training a single model to serve both roles could yield further performance gains.

During the ICT loop, we used a fixed held-out task set to identify the best-performing prompt at the end of each loop.
In real-world settings, such a repeatable evaluation set may not be available, necessitating alternative mechanisms for determining when to stop adaptation.
One such alternative is lifelong learning -- continuously updating the system prompt without a fixed stopping criterion -- though this raises open questions about the long-term stability of the generated prompts, which we leave for future work.\looseness-1

\section{Conclusion}
\label{sec:conclusion}
In this work, we investigated whether LLMs can learn from their own experience in-context in the realistic setting of online learning.
We introduced the In-context Training (ICT) task as an framework for evaluating cross-task self-improvement, in which an LLMs repeatedly  analyses batches of unseen episodes and generates system prompts intended to improve performance across a task family.
To train models for this setting, we proposed an RL-based fine-tuning approach that derives reward signals from replayed input tasks.
Across held-out task types, our trained reflectors consistently outperform an untrained baseline, showing that leaning from experience is itself a learnable capability. \looseness-1

We believe this has important implications for language agents.
Our work provides both an evaluation framework and a training methodology for cross-task self-learning, a capability likely to matter for general-purpose agents operating in open-ended environments.
We hope it shifts attention from reflection within a single task instance to reflection across tasks.
Because our approach requires no annotated examples and applies directly to agentic benchmarks, it is readily extensible. To support further research, we release MetaGym (see \Cref{sec:metagym}), a Python library for constructing meta-environments for LLM agentic tasks.

\newpage
\bibliographystyle{plainnat}
\bibliography{references}

\clearpage
\pagebreak
\appendix
\crefalias{section}{appendix}

\section{Reflector Prompt}
\label{appendix:prompting}

The reflector prompt is shared across both environments. The reflector receives the following system prompt:

\begin{tcolorbox}[breakable, fontupper=\small\ttfamily, title=Reflector system prompt]
You are an expert at analysing game-playing trajectories and improving agent instructions.

You will be given one or more episodes in which an agent attempts to complete a task guided by a system prompt. Your goal is to learn from these episodes to improve the agent's performance on future tasks.

Your job is to:\\
1. Identify what went well and what went wrong in the agent's behaviour.\\
2. Diagnose how the current system prompt contributed to those outcomes.\\
3. Write an improved system prompt that addresses the identified weaknesses to improve the agent's performance in future episodes.

Respond in EXACTLY this format --- no additional text outside these two sections:

ANALYSIS:\\
{[}Describe what succeeded, what failed, how the system prompt contributed to those outcomes, and what specific changes the improved prompt should make.{]}

IMPROVED PROMPT:\\
{[}The full improved system prompt, written to be used directly without modification.{]}
\end{tcolorbox}

The user message sent to the reflector follows this template:

\begin{tcolorbox}[breakable, fontupper=\small\ttfamily, title=Reflector user prompt template]
\#\# System Prompt Used

\{previous\_prompt\}

\#\# Episode Trajectories

=== Episode 1 ===\\
Success: Yes/No\\
\\
--- Step 1 ---\\
Observation: \{observation\}\\
Action: \{action\}\\
\\
--- Step 2 ---\\
Observation: \{observation\}\\
Action: \{action\}\\
\\
...\\
\\
Step limit reached. \textit{(only if unsuccessful)}\\
Total reward: \{reward\}\\
\\
\\
=== Episode 2 ===\\
Success: Yes/No\\
\\
--- Step 1 ---\\
Observation: \{observation\}\\
Action: \{action\}\\
\\
--- Step 2 ---\\
Observation: \{observation\}\\
Action: \{action\}\\
\\
...\\
\\
Step limit reached. \textit{(only if unsuccessful)}\\
Total reward: \{reward\}\\
\\

Analyse the trajectories and provide an improved system prompt.
\end{tcolorbox}

\section{Compute and Training Details}
\label{appendix:hyperparameters}

All experiments were conducted on a cluster node equipped with four NVIDIA GH200 Grace Hopper Superchips, each with 96 GB of GPU memory. Each reflector was trained for approximately 20 hours. Each ICT evaluation experiment, consisting of 10 full ICT loops, took approximately 2 node hours. In total, training and evaluation across all experiments required approximately 168 node hours.

\Cref{tab:hyperparams} lists the hyperparameters used to train the four reflectors. All runs use Qwen2.5-7B-Instruct as the base model, GRPO~\citep{shao2024deepseekmath} as the training algorithm, and the \textit{verl} framework~\citep{sheng2024hybridflow} for RL-based fine-tuning. Parameters that differ across runs reflect the longer in-context histories produced by different batch sizes $k$ and different observation lengths in different benchmarks. The checkpoint used for evaluation was selected based on validation performance.

\begin{table}[h]
\centering
\caption{Hyperparameters for reflector training across all four settings.}
\label{tab:hyperparams}
\small
\begin{tabular}{lcccc}
\toprule
\textbf{Hyperparameter} & \textbf{ALFWorld $k{=}3$} & \textbf{ALFWorld $k{=}5$} & \textbf{MiniHack $k{=}3$} & \textbf{MiniHack $k{=}5$} \\
\midrule
Base model & \multicolumn{4}{c}{Qwen2.5-7B-Instruct} \\
Algorithm & \multicolumn{4}{c}{GRPO} \\
Learning rate & \multicolumn{4}{c}{$3\times10^{-6}$} \\
Train batch size & \multicolumn{4}{c}{16} \\
PPO mini-batch size & \multicolumn{4}{c}{16} \\
PPO epochs & \multicolumn{4}{c}{2} \\
Rollout samples ($n$) & \multicolumn{4}{c}{8} \\
KL loss coefficient & \multicolumn{4}{c}{0.001} \\
\midrule
Max prompt length & 8192 & 8192 & 9216 & 10240 \\
Max response length & 2048 & 2048 & 2048 & 1024 \\
Rollout max model len & 10240 & 10240 & 11264 & 11264 \\
Checkpoint used & 160 & 120 & 70 & 60 \\
\bottomrule
\end{tabular}
\end{table}

\section{MetaGym}
\label{sec:metagym}
MetaGym is a Python library designed to support prompt optimisation and self-improvement research for LLM agents.
It natively implements the ICT task from the perspective of the reflector as a meta-agent.
Each meta-action is a system prompt, and each returned meta-observation is a batch of rollouts generated by the actor conditioned on that prompt.
Internally, the meta-environment spawns a batch of task instances and executes them in parallel using a chosen actor, providing native infrastructure for running the ICT loop described in \cref{sec:ict}.
MetaGym wraps a Gym-compatible environment \cite{1606.01540} inside a Gym-like meta-environment and handles the parallel execution and trajectory collection required by the ICT framework.
In our experiments, all task execution across ALFWorld and MiniHack was managed through MetaGym. The library ships with built-in support for both benchmarks and provides instructions for integrating new environments, with the aim of supporting future research on prompt optimisation and self-improvement in LLM agents. MetaGym is released publicly alongside this work at \url{https://github.com/yuvalshalev1/MetaGym}.

The following snippet illustrates the MetaGym API using ALFWorld:

\begin{tcolorbox}[breakable, fontupper=\small\ttfamily, title=MetaGym usage example]
from metagym import Actor\\
from metagym.envs.alfworld import ALFWorldMetaEnv\\
from metagym.reflector import Reflector\\
\\
\# Actor auto-launches a vLLM server for the model and shuts it down on exit\\
actor = Actor("Qwen/Qwen2.5-7B-Instruct")\\
\\
\# Initialise the meta-environment\\
env = ALFWorldMetaEnv(num\_envs=3, actor=actor, split="train", seed=42)\\
trajectories, info = env.reset()\\
\\
\# Reflector: analyses trajectories and rewrites the system prompt\\
reflector = Reflector("Qwen/Qwen2.5-7B-Instruct", base\_url=actor.base\_url)\\
\\
current\_prompt = "You are a helpful agent..."\\
for step in range(4):\\
\hspace*{2em}trajectories, reward, done, info = env.step(current\_prompt)\\
\hspace*{2em}print(f"Step \{step\}  win rate: \{reward:.0\%\}")\\
\hspace*{2em}reflection = reflector.reflect(trajectories, current\_prompt)\\
\hspace*{2em}print(reflection.analysis)\\
\hspace*{2em}current\_prompt = reflection.improved\_prompt\\
\\
\# Inspect a trajectory\\
trajectories[0].pretty\_print()
\end{tcolorbox}

\section{ALFWorld Details}
\label{appendix:alfworld}

For the ICT loop, we use the built-in training and evaluation splits provided by the ALFWorld benchmark. The reflector receives task batches sampled from the training split, while the full validation split is used for evaluation. Each episode is limited to 25 steps.

\subsection{Actor prompt}

The actor is implemented as a ReAct-style agent~\citep{yao2022react}.
The initial system prompt $sp_0$ is shared across all ALFWorld experiments. It is as follows:

\begin{tcolorbox}[breakable, fontupper=\small\ttfamily, title=ALFWorld actor system prompt]
You are an intelligent agent playing ALFWorld, a text-based game where you control a robot in a household environment.

Your goal is to complete tasks by interacting with objects in the environment.

You must use the ReACT (Reasoning and Acting) approach:\\
1. THINK about what you observe and what you should do next\\
2. Take an ACTION from the available actions\\
3. Observe the result and repeat

Format your responses EXACTLY as follows:\\
Thought: {[}your reasoning about the current situation and what to do{]}\\
Action: {[}exact action from the available actions list{]}
\end{tcolorbox}

Each turn, the actor receives the following user message, appended to the conversation history of all previous steps in the episode:

\begin{tcolorbox}[breakable, fontupper=\small\ttfamily, title=ALFWorld actor user prompt template]
\{observation\}

Available actions:\\
\{available\_actions\}

What is your next thought and action?
\end{tcolorbox}

\subsection{Trajectory example}

\begin{tcolorbox}[breakable, fontupper=\small\ttfamily, title=ALFWorld trajectory excerpt (pick\_two\_obj\_and\_place)]
Observation 0:\\
You are in the middle of a room. Looking quickly around you, you see a cabinet 4, a cabinet 3, a cabinet 2, a cabinet 1, a coffeemachine 1, a countertop 1, a drawer 8, a drawer 7, a drawer 6, a drawer 5, a drawer 4, a drawer 3, a drawer 2, a drawer 1, a fridge 1, a garbagecan 1, a microwave 1, a sidetable 1, a sinkbasin 1, a stoveburner 4, a stoveburner 3, a stoveburner 2, a stoveburner 1, and a toaster 1.\\
Your task is to: put two spoon in sidetable.\\
\\
Available actions: examine cabinet 1, go to cabinet 2, go to cabinet 3, go to cabinet 4, go to coffeemachine 1, go to countertop 1, go to drawer 1, go to drawer 2, go to drawer 3, go to drawer 4, go to drawer 5, go to drawer 6, go to drawer 7, go to drawer 8, go to fridge 1, go to garbagecan 1, go to microwave 1, go to sidetable 1, go to sinkbasin 1, go to stoveburner 1, go to stoveburner 2, go to stoveburner 3, go to stoveburner 4, go to toaster 1, help, inventory, look, open cabinet 1\\
Thought 1: The task is to put two spoons in the sidetable. Spoons are likely to be found in cabinets or drawers. I will start by checking the cabinets.\\
Action 1: go to cabinet 1\\
\\
Observation 1:\\
You arrive at cabinet 1. The cabinet 1 is closed.\\
\\
Available actions: close cabinet 1, examine cabinet 1, go to cabinet 2, go to cabinet 3, go to cabinet 4, go to coffeemachine 1, go to countertop 1, go to drawer 1, go to drawer 2, go to drawer 3, go to drawer 4, go to drawer 5, go to drawer 6, go to drawer 7, go to drawer 8, go to fridge 1, go to garbagecan 1, go to microwave 1, go to sidetable 1, go to sinkbasin 1, go to stoveburner 1, go to stoveburner 2, go to stoveburner 3, go to stoveburner 4, go to toaster 1, help, inventory, look, take mug 1 from cabinet 1, take peppershaker 1 from cabinet 1\\
Thought 2: Since the cabinet is closed, I need to open it to see if there are any spoons inside.\\
Action 2: open cabinet 1\\
\\
Observation 2:\\
You open the cabinet 1. The cabinet 1 is open. In it, you see a mug 1, and a peppershaker 1.\\
\\
Available actions: examine cabinet 2, go to cabinet 1, go to cabinet 3, go to cabinet 4, go to coffeemachine 1, go to countertop 1, go to drawer 1, go to drawer 2, go to drawer 3, go to drawer 4, go to drawer 5, go to drawer 6, go to drawer 7, go to drawer 8, go to fridge 1, go to garbagecan 1, go to microwave 1, go to sidetable 1, go to sinkbasin 1, go to stoveburner 1, go to stoveburner 2, go to stoveburner 3, go to stoveburner 4, go to toaster 1, help, inventory, look, open cabinet 2\\
Thought 3: The cabinet did not contain any spoons. I should check the other cabinets and drawers for spoons.\\
Action 3: go to cabinet 2
\end{tcolorbox}

\section{MiniHack Details}
\label{appendix:minihack}

MiniHack environments are procedurally generated, so task instances are produced by sampling random seeds. The reflector receives task batches sampled from a designated range of training seeds, while a fixed set of 32 seeds is held out and used consistently as the validation set for the ICT loop. Each episode is limited to 25 steps, with the exception of the $k{=}5$ training run, where this limit was reduced to 18 steps to keep episode trajectories within the model's context length.

MiniHack natively exposes a large number of actions, many of which are irrelevant to the tasks in our benchmark. To reduce ambiguity and keep the context concise, we always provide the actor with a curated list of available actions that are suitable for our task set, rather than the full action space.

\subsection{Actor prompt}

The actor is implemented as a ReAct-style agent~\citep{yao2022react}.
The initial system prompt $sp_0$ is shared across all MiniHack experiments. It is as follows:

\begin{tcolorbox}[breakable, fontupper=\small\ttfamily, title=MiniHack actor system prompt]
You are an intelligent agent playing MiniHack, a text-based dungeon exploration game.
You are placed in an ASCII-based grid world in which your character is represented by the ``@'' symbol. Explore the grid world and find your goal.

You must use the ReACT (Reasoning and Acting) approach:\\
1. THINK about what you observe and what you should do next\\
2. Take an ACTION from the available actions\\
3. Observe the result and repeat

Format your responses EXACTLY as follows:\\
Thought: {[}your reasoning about the current situation and what to do{]}\\
Action: {[}exact action from the available actions list{]}
\end{tcolorbox}

Each turn, the actor receives the following user message, appended to the conversation history of all previous steps in the episode:

\begin{tcolorbox}[breakable, fontupper=\small\ttfamily, title=MiniHack actor user prompt template]
Task: play \{task\}\\
\{observation\}

Available actions:\\
\{available\_actions\}

What is your next thought and action?
\end{tcolorbox}

\subsection{Trajectory example}

\begin{tcolorbox}[breakable, fontupper=\small\ttfamily, title=MiniHack trajectory excerpt (MiniHack-PutOn-v0)]
Observation 0:\\
Hello Agent, welcome to NetHack! You are a neutral human Caveman.\\
\\
\hspace*{2em}."...\\
\hspace*{2em}.....\\
\hspace*{2em}.....\\
\hspace*{2em}..@..\\
\hspace*{2em}.....\\
\\
Available actions: step n, step s, step e, step w, step ne, step nw, step se, step sw, run n, run s, run e, run w, run ne, run nw, run se, run sw, read, eat, puton, zap, wield, wear, pickup\\
Thought 1: The game has just started. I will start by moving north to explore.\\
Action 1: step n\\
\\
Observation 1:\\
\hspace*{2em}."...\\
\hspace*{2em}.....\\
\hspace*{2em}..@..\\
\hspace*{2em}..<..\\
\hspace*{2em}.....\\
\\
Available actions: step n, step s, step e, step w, step ne, step nw, step se, step sw, run n, run s, run e, run w, run ne, run nw, run se, run sw, read, eat, puton, zap, wield, wear, pickup\\
Thought 2: I can see a "<" symbol. I will move east to investigate further.\\
Action 2: step e\\
\\
Observation 2:\\
\hspace*{2em}."...\\
\hspace*{2em}.....\\
\hspace*{2em}...@.\\
\hspace*{2em}..<..\\
\hspace*{2em}.....\\
\\
Available actions: step n, step s, step e, step w, step ne, step nw, step se, step sw, run n, run s, run e, run w, run ne, run nw, run se, run sw, read, eat, puton, zap, wield, wear, pickup\\
Thought 3: Moving south to continue exploring.\\
Action 3: step s
\end{tcolorbox}

\end{document}